\documentclass[pdflatex,sn-mathphys-num]{sn-jnl}

\usepackage{graphicx}%
\usepackage{multirow}%
\usepackage{amsmath,amssymb,amsfonts}%
\usepackage{amsthm}%
\usepackage{mathrsfs}%
\usepackage[title]{appendix}%
\usepackage{xcolor}%
\usepackage{textcomp}%
\usepackage{manyfoot}%
\usepackage{booktabs}%
\usepackage{algorithm}%
\usepackage{algorithmicx}%
\usepackage{algpseudocode}%
\usepackage{listings}%
\usepackage{listings}
\usepackage{color}
\usepackage{xcolor}
\usepackage{graphicx}
\usepackage{float}

\definecolor{dkblue}{rgb}{0,0.39,0}
\definecolor{gray}{rgb}{0.66,0.66,0.66}
\definecolor{mauve}{rgb}{0.91,0.33,0.50}
\definecolor{gold}{rgb}{1,0.84,0}

\begin{document}

\title[Predicting Coronary Heart Disease Using a Suite of Machine Learning Models]{Predicting Coronary Heart Disease Using a Suite of Machine Learning Models}

\author[1,2]{\fnm{Jamal} \sur{Al-Karaki}}\email{Jamal.Al-Karaki@zu.ac.ae}

\author[3]{\fnm{Philip} \sur{Ilono}}

\author[3]{\fnm{Sanchit} \sur{Baweja}}

\author[3]{\fnm{Jalal} \sur{Naghiyev}}

\author[3]{\fnm{Raja Singh} \sur{Yadav}}

\author*[1]{\fnm{Muhammad Al-Zafar} \sur{Khan}}\email{Muhammad.Al-ZafarKhan@zu.ac.ae}

\affil*[1]{\orgname{College of Interdisciplinary Studies (CIS), Zayed University}, \state{Abu Dhabi}, \country{UAE}}

\affil[2]{\orgdiv{College of Engineering}, \orgname{The Hashemite University}, \city{Zarqa}, \country{Jordan}}

\affil[3]{Independent Researcher}

\abstract{Coronary Heart Disease affects millions of people worldwide and is a well-studied area of healthcare. There are many viable and accurate methods for the diagnosis and prediction of heart disease, but they have limiting points such as invasiveness, late detection, or cost. Supervised learning via machine learning algorithms presents a low-cost (computationally speaking), non-invasive solution that can be a precursor for early diagnosis. In this study, we applied several well-known methods and benchmarked their performance against each other. It was found that Random Forest with oversampling of the predictor variable produced the highest accuracy of 84\%.
}


\maketitle

\section{Introduction}

Predicting coronary heart disease (CHD) remains a significant challenge in medical diagnostics, with ongoing research aimed at improving predictive accuracy. This literature review examines machine learning techniques for CHD prediction, focusing on methodologies, results, and limitations of existing research. It also explores how these techniques utilize both labeled and unlabeled data to enhance model performance and address issues related to data availability.

Coronary Heart Disease (CHD) is caused by the blockage of blood flow to the heart through the constriction of coronary arteries. Such blockages are typically a consequence of the accumulation of fatty deposits within the arteries. CHD was reported to have roughly 110 million cases worldwide in 2015, and cardiovascular diseases, in general, accounted for roughly 17.9 million fatalities in the same year \cite{roth2017global}. Early diagnosis is important as failure to detect early on could lead to heart attacks or cardiac arrest. Identification also enhances the effectiveness of varied treatment options such as medication, lifestyle changes through exercise and diet or even surgical operations

Imaging techniques such as Coronary Computed Tomography Angiography (CCTA) render 3D images to visualize the internal structure of coronary arteries. Such observations aid in identifying coronary artery disease and severe stenosis, and this process is also non-invasive \cite{fairbairn2018real}. Fractional Flow Reserve (FFR) is minimally invasive and measures the fluctuations in pressure across the coronary artery to establish the hemodynamic significance. Quantitative Coronary Angiography (QCA) is an alternative to FFR, and both tend to be compared to assess lesions. QCA is invasive and measures the diameter of the stenosis to benchmark against reference measurements \cite{fairbairn2018real}. 

Other methods include stress testing, cardiac biomarkers measurements through blood tests, and clinical risk assessment through diamond-forrester models are other viable tests for coronary heart disease diagnosis. The main limitations of such methods are that some are invasive, and the generalization and accuracy of the non-invasive methods are improvable as CCTA is still limited for predicting physiologically significant coronary artery disease.

While we acknowledge that none of the methods applied are new, there does not exist a comprehensive study that benchmarks so many different models against each other, and thus, we believe that this research, and our original contribution in this regard, serves as a primer into the field. Further, all the results obtained are highly reproducible and require simple, non-abstract code. Lastly, we would like to point out that the methods applied here can easily be applied to any other dataset, and the benchmarking methods in a neat tabular format can be adopted. 

\noindent This paper is organized as follows:

In Sec. \ref{lit rev}, we peruse some of the contemporary studies that applied Machine Learning (ML) and Deep Learning (DL) to data.

In Sec. \ref{theory}, we outline the mathematical details underpinning the models we have applied and describe their calculational mechanics.

In Sec. \ref{experiments}, we describe the process undertaken to clean the data and thereafter present the results.

In Sec. \ref{conclusion}, we reflect upon the results obtained and provide details of future studies that can offshoot from this one.

\section{Literature Review}\label{lit rev}

The study in \cite{inbook} focuses on developing a predictive model for coronary heart disease risk using machine learning techniques. The authors employed the CRISP-DM (Cross-Industry Standard Process for Data Mining) framework for their methodology. Among various models tested, the Decision Tree algorithm proved to be the most effective, achieving notable performance metrics: an accuracy of 0.884, an AUC value of 0.942, and an $F1$ score of 0.881. Cross-validation was utilized as the sampling method to ensure a robust evaluation of the model's performance.

Authors in \cite{krishnani2019prediction} evaluated the effectiveness of Random Forest, Decision Trees, and $K$-Nearest Neighbors algorithms in predicting Coronary Heart Disease using the ``Framingham Heart Study'' dataset. By preprocessing the data through techniques such as standardization and normalization and employing $K$-fold cross-validation, the research demonstrates that Random Forest excels in prediction accuracy compared to the other algorithms.

The study in \cite{gonsalves2019prediction} explores the effectiveness of machine learning techniques, specifically Naive Bayes, SVMs, and Decision Trees, in predicting coronary heart disease. Using a dataset of 462 instances from the South African Heart Disease dataset and employing 10-fold cross-validation, the research demonstrates the potential of these methods for disease prediction. However, the study is limited by its focus on only three machine learning techniques and the relatively small size of the dataset, which may affect the generalizability of the findings.

 In \cite{bemando2021machine}, the authors compare Gaussian Naive Bayes, Bernoulli Naive Bayes, and Random Forest algorithms in predicting coronary heart disease. Using the Cleveland dataset from the UCI repository, the study evaluates these models based on accuracy, precision, $F1$ score, and recall. The results demonstrate that both Gaussian and Bernoulli Naive Bayes algorithms outperform Random Forest, suggesting their superior effectiveness in coronary heart disease prediction.

 Authors of study \cite{sk2023coronary} introduced a hybrid approach integrating Decision Tree and AdaBoost algorithms for predicting and classifying coronary heart disease. The methodology leverages the strengths of both algorithms to enhance prediction accuracy and classification performance. Evaluated through metrics like accuracy, True Positive Rate, and Specificity, this hybrid method demonstrates improved efficacy in CHD prediction compared to traditional approaches.

The study in \cite{miao2018coronary} focuses on diagnosing coronary heart disease using a deep neural network (DNN) with a multilayer perceptron architecture. It employs regularization and dropout techniques to enhance model performance, specifically designed for both the classification of training data and the prediction of new patient cases. The model was trained and evaluated using a dataset of 303 clinical instances from the Cleveland Clinic Foundation.

The paper \cite{satu2018exploring} investigates the effectiveness of semi-supervised learning algorithms in identifying key predictors of coronary heart disease. Utilizing Cleveland and Hungarian heart disease datasets, the study applies Collective Wrapper, Filtered Collective, and Yet Another Semi-Supervised Idea algorithms to subsets of varying sizes (33\%, 65\%, and 100\%). Performance is assessed through accuracy, $F1$ score, and ROC curve metrics. By sequentially removing attributes, the research distinguishes between significant and irrelevant factors, providing insights into the factors influencing heart disease prediction. 

Authors \cite{lin2023coronary} presented a novel adversarial domain-adaptive multichannel graph convolutional network (DAMGCN) for predicting coronary heart disease. This approach integrates a two-channel graph convolutional network with local and global consistency for feature aggregation alongside an attention mechanism to unify node representations across different graphs. A domain adversarial module is employed to minimize discrepancies between source and target domain classifiers and optimize three loss functions for effective cross-domain knowledge transfer. The proposed DAMGCN model aims to overcome the limitation of existing GNN models in transferring knowledge across diverse datasets, a significant challenge in CHD research.

The study in \cite{malik2023coronary} presents a hybrid model integrating Gaussian Fuzzy C-Means Clustering (GKFCM) with a Recurrent Neural Network for predicting coronary heart disease. The methodology involves normalizing the dataset, applying GKFCM for clustering, and then leveraging the combined GKFCM-RNN approach to achieve a high prediction accuracy of 99\%. This innovative approach demonstrates significant promise in enhancing predictive accuracy for coronary heart disease through advanced ML techniques.

\section{Theory}\label{theory}
In this section, we present the theory of all the ML models used in this study. 

Suppose that our dataset $\mathcal{D}=\left\{\left(\mathbf{x}_{i},y_{i}\right)\right\}_{i}\in\mathbb{R}^{m\times n}$ composed of $m$ examples and $n$ features $\mathbf{x}=\left(x_{1},x_{2},\ldots,x_{n}\right)$, with labels $y$. For each of the models below, under the supervised learning paradigm, we seek to find a function $f:\mathbf{x}\to y$ that maps the features to the label.

\subsection{Logistic Regression}
With this model, we aim to predict the probability of binary outcomes given by
\begin{equation}
p(y=1|\mathbf{x})=\frac{1}{1+\exp\left[-\left(w_{0}+w_{1}x_{1}+\ldots+w_{n}x_{n}\right)\right]}=\sigma(\mathbf{w}^{T}\mathbf{x}+b),
\end{equation}
where $\mathbf{w}=\left(w_{1},w_{2},\ldots,w_{n}\right)$ are the weights, for $0\leq i\leq 1$, $b=w_{0}$ is the bias term, and $\sigma(\mathcal{X})=1/\left(1+e^{-\mathcal{X}}\right)$ is the sigmoid activation function for arbitrary $\mathcal{X}$. 

\subsection{Linear Discriminant Analysis}
In Linear Discriminant Analysis (LDA), each class of the predictor variables is assumed to be drawn from a Gaussian distribution, and Bayes theorem is used to classify them based on the posterior probability. The discriminant function for class $c\in\mathcal{C}$ is
\begin{equation}
\delta_{c}(\mathbf{x})=\mathbf{x}^{T}\boldsymbol{\Sigma}^{-1}\boldsymbol{\mu}_{c}-\frac{1}{2}\boldsymbol{\mu}_{c}^{T}\boldsymbol{\Sigma}^{-1}\boldsymbol{\mu}_{c}+\log\pi_{c},
\end{equation}
where $\boldsymbol{\Sigma}$ is the covariance matrix, $\boldsymbol{\mu}_{c}$ is the mean in class $c$ and $\pi_{c}$ is the prior probability of class $c$. The class is then assigned according to the decision rule
\begin{equation}
\hat{y}=\underset{c\in\mathcal{C}}{\arg\max}\;\delta_{c}(\mathbf{x}).
\end{equation}

\subsection{Support Vector Machines}
The Support Vector Machine (SVM) model seeks to find the optimal separating hyperplane between classes that maximizes the margin (distance between the planes with equation $\mathbf{w}^{T}\mathbf{x}=-1$ and $\mathbf{w}^{T}\mathbf{x}=1$). Mathematically, this is expressed as the optimization problem
\begin{equation}
\begin{aligned}
&\underset{\mathbf{w},b}{\min}\;\mathcal{L}=\underset{\mathbf{w},b}{\min}\;\frac{1}{2}\mathbf{w}^{T}\mathbf{w} \\
\text{subject to:}\;\;\;\;\;&y_{i}\left(\mathbf{w}^{T}\mathbf{x}_{i}+b\right)\geq 1\quad\forall i.
\end{aligned}
\end{equation}

\subsection{Decision Trees}
Decision Trees seek to recursively split the data at each node based on a feature and threshold to maximize information gain or minimize impurity. The impurity can be either the Gini value or the entropy. Mathematically, the Gini and entropy, respectively, are
\begin{equation}
\begin{aligned}
G=&\;1-\sum_{c\in\mathcal{C}}p_{c}^{2}, \\
S=&\;-\sum_{c\in\mathcal{C}}p_{c}\log p_{c},    
\end{aligned}
\end{equation}
where $p_{c}$ is the probability of samples belonging to class $c$ at a particular node.

\subsection{Random Forest}
Random forest is an ensemble learning method that operates by constructing a large number of decision trees during training and then outputting the mode of the classes of the individual trees.

The Random Forest algorithm combines the output of multiple decision trees to provide more accurate and stable predictions. Algorithmically, it works as follows:
\begin{enumerate}
\item \textbf{Bagging/Bootstrap Aggregating:} From the original dataset $\mathcal{D}$, a large number, say $B$, of subsets $\mathcal{D}_{b}$ are created by randomly sampling with replacement. Each of these subsets will serve as the training dataset for a specific decision tree $T_{b}$. 
\item \textbf{Construction of Trees:} For each subset, a decision tree is built. However, unlike traditional decision trees, each node in the tree is split based on a random subset of the features. This is another key to the success of the Random Forest algorithm, as it helps to reduce the correlation between the trees.
\item \textbf{Classification:} Each decision tree makes a prediction $h_{b}(x_{\text{test}})$ for instance $x_{\text{test}}$, and the random forest outputs the class that receives the majority vote
\begin{equation}
\hat{y}=\text{mode}\left[h_{1}(x_{\text{test}}),h_{2}(x_{\text{test}}),\ldots,h_{B}(x_{\text{test}})\right],
\end{equation}
where $\text{mode}$ is the most frequent class label among the predictions.
\end{enumerate}

\subsection{Naive Bayes}
The Naive Bayes model is probabilistic in nature and uses Bayes Theorem from probability theory to perform classification. It assumes that the features in the dataset are conditionally independent given the class label.

The goal is to estimate the probability of a class $c\in\mathcal{C}$ given a feature vector $\mathbf{x}=\left(x_{1},x_{2},\ldots,x_{n}\right)$ of observed features. The probability of class $c$ is given by
\begin{equation}
p(c|\mathbf{x})=\frac{p(\mathbf{x}|c)p(c)}{p(\mathbf{x})},
\end{equation}
where $p(c|\mathbf{x})$ is the posterior probability, $p(\mathbf{x}|c)$ is the likelihood, $p(c)$ is the prior probability, and $p(\mathbf{x})$ is the evidence. The predicted class is then given by
\begin{equation}
\hat{c}=\underset{c\in\mathcal{C}}{\arg\max}\;p(c|\mathbf{x}).
\end{equation}

\subsection{$K$-Nearest Neighbors}
The $K$-Nearest Neighbors ($K$-NN) algorithm is a non-parametric, makes no assumptions about the underlying data distribution, and lazy learning, does not learn a model during the training phase; instead, it stores the training data and uses it directly during prediction, algorithm.

The idea is that similar data points will likely exist near one another. When making a prediction, $K$-NN looks at the $K$ closest training examples in the feature space and assigns a label based on the majority, for classification tasks, of those neighbors. Typically, the distance between the new datapoint and all the datapoints in the training set is calculated, and $K$ closest datapoints to the new datapoint based on the computed distance are chosen. A typical metric used is the $\ell^{2}$ distance 
\begin{equation}
d(x_{i},x_{j})=\sqrt{\sum_{n}\left(x_{i}^{n}-x_{j}^{n}\right)^{2}}.
\end{equation}
If $K$ is too small, the model becomes sensitive to noise and leads to overfitting, and coversely if $k$ is too large, the model becomes becomes too generalized and leads to overfitting. 

\subsection{Extreme Gradient Boosting}
Extreme Gradient Boosting (XGBoost) is an ensemble learning method which combines the predictions of multiple weak learners, typically decision trees, to produce a strong learner. The model works by sequentially adding trees, where each subsequent tree tries to correct the residuals (errors) of the previous trees. In order to compensate for overfitting, the model includes $L_{1}$ (lasso) and $L_{2}$ (ridge) regression terms. 

Mathematically, the model works as follows: 
\begin{enumerate}
\item The initial prediction is calculated using 
\begin{equation}
\hat{y}_{i}^{\left(0\right)}=\frac{1}{m}\sum_{i=1}^{m}y_{i}.
\end{equation}
\item At each iteration, a set of trees $t\in\mathcal{T}$ is trained according to
\begin{equation}
\hat{y}_{i}^{\left(t+1\right)}=\hat{y}_{i}^{\left(t\right)}+f_{t}(x_{i}),
\end{equation}
where $f_{t}(x_{i})$ is the prediction of the $t^{\text{th}}$ tree.
\item The objection function
\begin{equation}
\mathcal{L}^{\left(t\right)}=\sum_{i=1}^{m}J(y_{i},y_{i}^{\left(t\right)})+\sum_{k=1}^{y}\Omega(f_{k}),
\end{equation}
where $J$ is the loss function between the true and target labels -- usually chosen to be log-loss in classification tasks such as the problem we are solving -- and $\Omega$ is the penalization term that regulates the model complexity, given by
\begin{equation}
\Omega(f_{k})=\gamma T+\frac{\lambda}{2}\sum_{j}||w_{j}||_{2}^{2},    
\end{equation}
with $0<\gamma<1$ is the control parameter, $T$ is the number of leaves in the tree, and $0<\lambda<1$ is the $L_{2}$ control parameter. 
\item Each tree is constructed in a greedy manner by selecting splits that minimize the loss function. XGBoost uses the gain of each split to determine the quality of a split, which is calculated as
\begin{equation}
\text{Gain}=\frac{1}{2}\left[\frac{G_{L}^{2}}{H_{L}+\lambda}+\frac{G_{R}^{2}}{H_{R}+\lambda}-\frac{\left(G_{L}+G_{R}\right)^{2}}{H_{L}+H_{R}+\lambda}\right]-\gamma,
\end{equation}
where $G_{L}$ and $G_{R}$ are the sums of the gradients for the left and right child nodes, respectively, and $H_{L}$ and $H_{R}$ are the sums of the Hessians for the left and right child nodes.
\end{enumerate}

\section{Experiments}\label{experiments}
The dataset consists of $16$ fields and $4\;240$ examples, of which there are 645 missing values in total. In Fig. \ref{fig1}, we represent the distribution of missing values per feature. Since the missing values represented only a small proportion of the overall size of the dataset, they were dropped and the resulting dataset contains $3\;658$ examples. The data was normalized using min-max scaling in order to bring them the features in the non-categorical types to the range $\left[0,1\right]$ according to: For feature $x_{i}\in\mathbf{x}$, the normalized value of the feature is given by
\begin{equation}
x_{\text{norm}}=\frac{x-x_{\min}}{x_{\max}-x_{\min}},
\end{equation}
where $x_{\min}$ is the minimum value of the feature, and $x_{\max}$ is the maximum value of the feature. 

\begin{figure}[htpb]
    \centering
    \includegraphics[width=1\linewidth]{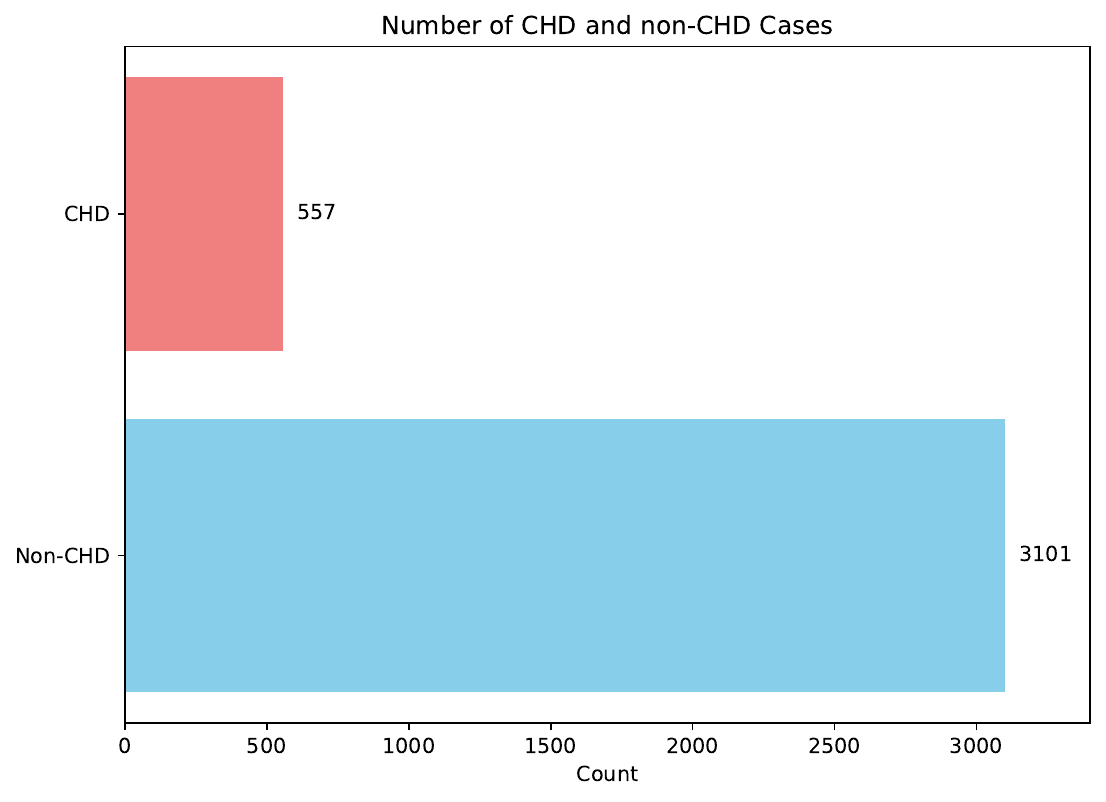}
    \caption{Counts of the classes in the TenYearCHD variable}
    \label{fig3}
\end{figure}

\begin{figure}[htpb]
\centering
\includegraphics[width=1.2\linewidth]{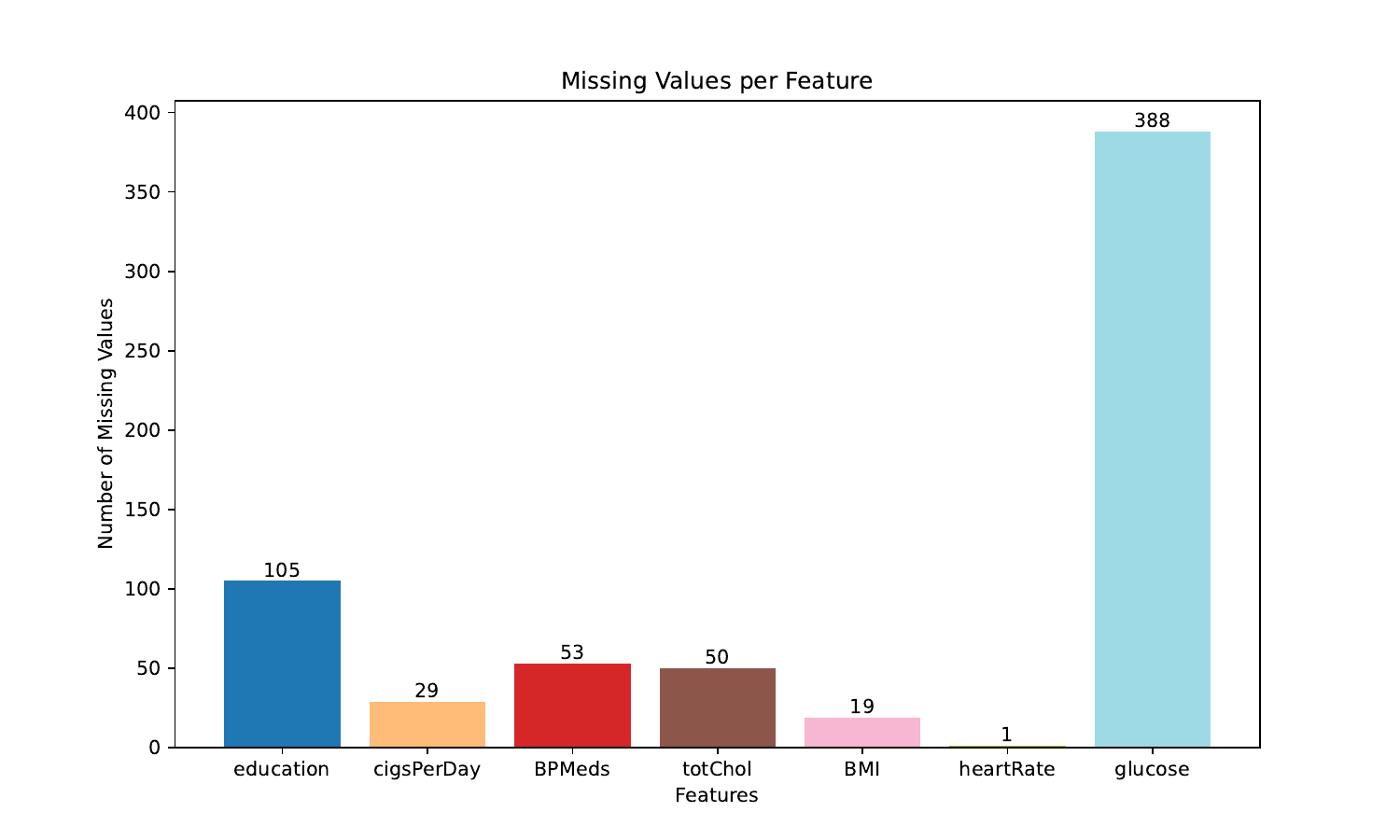}
\caption{Missing values per feature. Only those features that have missing values are depicted.}
\label{fig1}
\end{figure}

Further, we observe that a severe class imbalance occurs in the predictor variable \textbf{TenYearCHD}: $3\;101$ for CHD cases, and $557$ for non-CHD cases; see Fig. \ref{fig3} for a graphical depiction.  

Thus, we resort to balance the class balancing via random undersampling and random oversampling. 

The data was then split into training data ($70\%$) and testing/validation data ($30\%$), whereupon the eight different model types were built. The results of the model training together with the evaluation metrics are summarized in Tabs. \ref{tab1} and \ref{tab2}. 

\begin{table}[!htpb]
\centering
\begin{tabular}{lcccc}
\hline
&\textbf{Precision:}$P=\frac{TP}{TP+FP}$ &\textbf{Recall:}$R=\frac{TP}{TP+FN}$ &$\boldsymbol{F1:}F1=\frac{2\times P\times R}{P+R}$ \\
\hline
\textbf{Logistic Regression} & & & & \\ 
Non-CHD &0.91 &0.67 &0.77 \\
CHD &0.29 &0.66 &0.40 \\
Accuracy &$-$ &$-$ &$\boxed{0.67}$ \\
Macro Average &0.60 &0.67 &0.59 \\
Weighted Average &0.81 &0.67 &0.71 \\
\hline
\textbf{Linear Discriminant Analysis} & & & & \\ 
Non-CHD &0.91 &0.69 &0.78 \\
CHD &0.30 &0.66 &0.41 \\
Accuracy &$-$ &$-$ &$\boxed{0.68}$ \\
Macro Average &0.60 &0.67 &0.60 \\
Weighted Average &0.81 &0.68 &0.72 \\
\hline 
\textbf{Support Vector Machines} & & & & \\ 
Non-CHD &0.91 &0.66 &0.76 \\
CHD &0.28 &0.66 &0.39 \\
Accuracy &$-$ &$-$ &$\boxed{0.66}$ \\
Macro Average &0.59 &0.66 &0.58 \\
Weighted Average &0.80 &0.66 &0.70 \\
\hline 
\textbf{Random Forest} & & & & \\ 
Non-CHD &0.91 &0.67 &0.77 \\
CHD &0.28 &0.66 &0.40 \\
Accuracy &$-$ &$-$ &$\boxed{0.67}$ \\
Macro Average &0.59 &0.66 &0.58 \\
Weighted Average &0.80 &0.67 &0.71 \\
\hline 
\textbf{Decision Trees} & & & & \\ 
Non-CHD &0.86 &0.58 &0.69 \\
CHD &0.20 &0.53 &0.29 \\
Accuracy &$-$ &$-$ &$\boxed{0.57}$ \\
Macro Average &0.53 &0.56 &0.49 \\
Weighted Average &0.75 &0.57 &0.63 \\
\hline 
\textbf{Naive Bayes} & & & & \\ 
Non-CHD &0.87 &0.90 &0.88 \\
CHD &0.40 &0.34 &0.36 \\
Accuracy &$-$ &$-$ &$\boxed{0.80}$ \\
Macro Average &0.63 &0.62 &0.62 \\
Weighted Average &0.79 &0.80 &0.80 \\
\hline
\textbf{$K$-Nearest Neighbor} & & & & \\ 
Non-CHD &0.88 &0.66 &0.75 \\
CHD &0.24 &0.54 &0.33 \\
Accuracy &$-$ &$-$ &$\boxed{0.64}$ \\
Macro Average &0.56 &0.60 &0.54 \\
Weighted Average &0.77 &0.64 &0.68 \\
\hline 
\textbf{XGBoost} & & & & \\ 
Non-CHD &0.90 &0.68 &0.77 \\
CHD &0.28 &0.63 &0.39 \\
Accuracy &$-$ &$-$ &$\boxed{0.67}$ \\
Macro Average &0.59 &0.65 &0.58 \\
Weighted Average &0.80 &0.67 &0.71 \\
\hline
\end{tabular}
\caption{Metrics of the various ML models applied using undersampling. $TP$ represents true positives (the number of correct positive predictions), $FP$ represents false positives (the number of incorrect positive predictions), $FN$ represents false negatives (the number of positive instances that were incorrectly predicted as negative), and the support is the number of actual occurrences of each class in the dataset.}
\label{tab1}
\end{table}

\begin{table}[!htpb]
\centering
\begin{tabular}{lcccc}
\hline
&\textbf{Precision} &\textbf{Recall} &$\boldsymbol{F1}$ \\
\hline
\textbf{Logistic Regression} & & & & \\ 
Non-CHD &0.90 &0.71 &0.80 \\
CHD &0.30 &0.62 &0.41 \\
Accuracy &$-$ &$-$ &$\boxed{0.70}$ \\
Macro Average &0.60 &0.67 &0.60 \\
Weighted Average &0.80 &0.70 &0.73 \\
\hline
\textbf{Linear Discriminant Analysis} & & & & \\ 
Non-CHD &0.91 &0.71 &0.79 \\
CHD &0.30 &0.63 &0.41 \\
Accuracy &$-$ &$-$ &$\boxed{0.70}$ \\
Macro Average &0.60 &0.67 &0.60 \\
Weighted Average &0.81 &0.70 &0.73 \\
\hline 
\textbf{Support Vector Machines} & & & & \\ 
Non-CHD &0.90 &0.67 &0.77 \\
CHD &0.28 &0.65 &0.39 \\
Accuracy &$-$ &$-$ &$\boxed{0.67}$ \\
Macro Average &0.59 &0.66 &0.58 \\
Weighted Average &0.80 &0.67 &0.71 \\
\hline 
\textbf{Random Forest} & & & & \\ 
Non-CHD &0.85 &0.98 &0.91 \\
CHD &0.53 &0.13 &0.21 \\
Accuracy &$-$ &$-$ &$\boxed{0.84}$ \\
Macro Average &0.69 &0.55 &0.56 \\
Weighted Average &0.80 &0.84 &0.79 \\
\hline 
\textbf{Decision Trees} & & & & \\ 
Non-CHD &0.86 &0.89 &0.87 \\
CHD &0.31 &0.25 &0.27 \\
Accuracy &$-$ &$-$ &$\boxed{0.78}$ \\
Macro Average &0.58 &0.57 &0.57 \\
Weighted Average &0.76 &0.78 &0.77 \\
\hline 
\textbf{Naive Bayes} & & & & \\ 
Non-CHD &0.87 &0.91 &0.89 \\
CHD &0.42 &0.31 &0.36 \\
Accuracy &$-$ &$-$ &$\boxed{0.81}$ \\
Macro Average &0.65 &0.61 &0.62 \\
Weighted Average &0.79 &0.81 &0.80 \\
\hline
\textbf{$K$-Nearest Neighbor} & & & & \\ 
Non-CHD &0.86 &0.71 &0.78 \\
CHD &0.23 &0.43 &0.0.30 \\
Accuracy &$-$ &$-$ &$\boxed{0.67}$ \\
Macro Average &0.55 &0.57 &0.54 \\
Weighted Average &0.76 &0.67 &0.70 \\
\hline 
\textbf{XGBoost} & & & & \\ 
Non-CHD &0.88 &0.74 &0.80 \\
CHD &0.28 &0.50 &0.36 \\
Accuracy &$-$ &$-$ &$\boxed{0.70}$ \\
Macro Average &0.58 &0.62 &0.58 \\
Weighted Average &0.78 &0.70 &0.73 \\
\hline
\end{tabular}
\caption{Metrics of the various ML models applied using overersampling.}
\label{tab2}
\end{table}

\begin{figure}[htpb]
\centering
\includegraphics[width=1.2\linewidth]{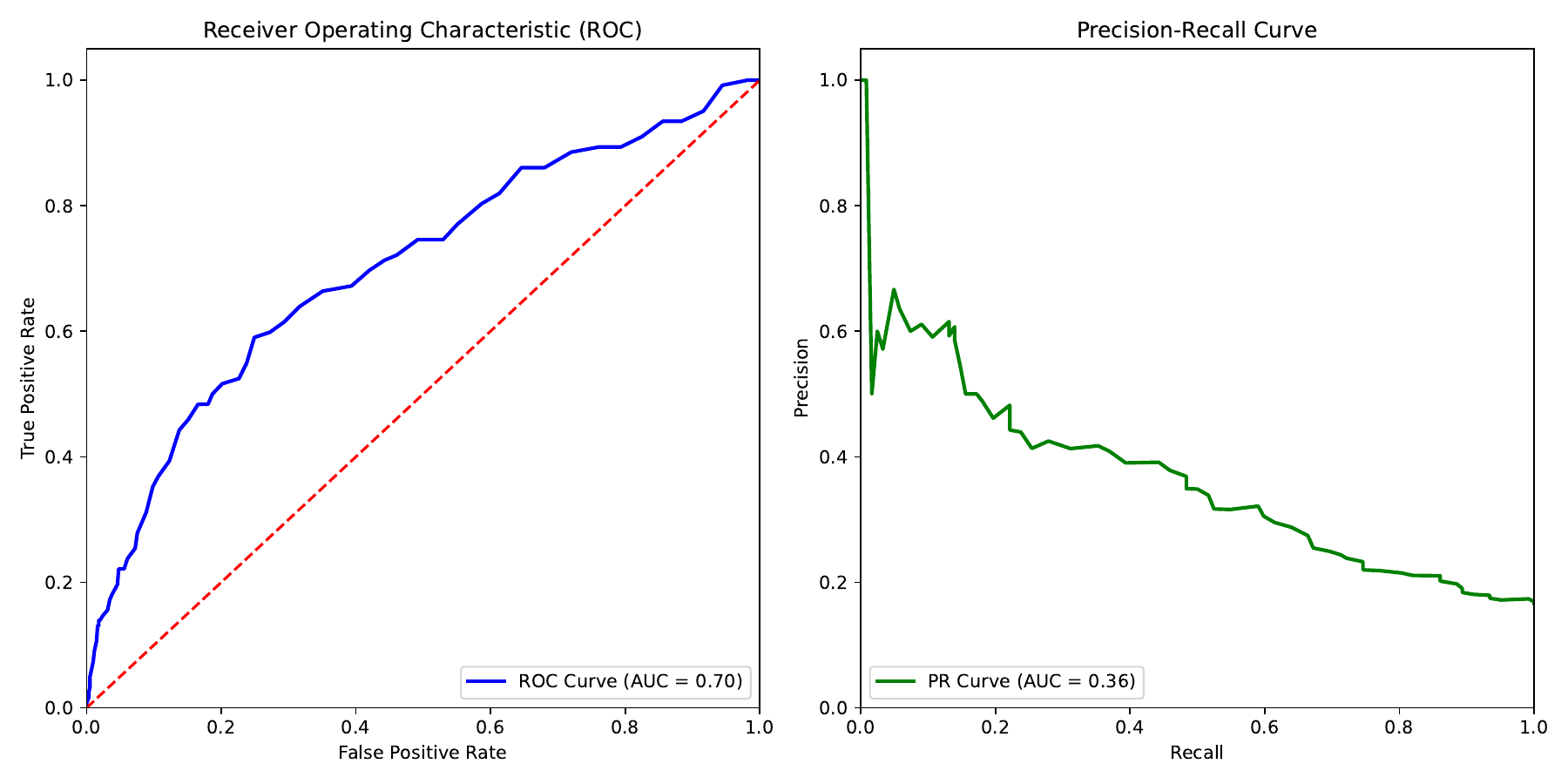}
\caption{\textbf{Left:} ROC. \textbf{Right:} Precision-Recall curve for the random forest model with oversampling.}
\label{fig4}
\end{figure}

From Tabs. \ref{tab1} and \ref{tab2}, we observe that overall random oversampling produced better results than random undersampling. Furthermore, the best performing model, in terms of accuracy, was the random forest model with an accuracy of $84\%$. In Fig. \ref{fig4}, we show the behavior of the receiver operating curve (ROC) and the precision-recall curve. The area under the ROC (AUROC) is which indicates that the model's predictive power is above random chance ($\text{AUC}=0.5$), but it is far from perfect. The model's ROC curve shows a fairly smooth ascent, but it's clear there is room for improvement, especially in reducing false positives. The AUC for Precision-Recall is 0.36, which is quite low. This suggests that the model struggles with maintaining high precision at high recall, meaning the model might produce many false positives when it tries to capture all the actual positives. The steep drop in precision as recall increases indicates that the model initially performs well at predicting positives, but as the recall grows, the model starts misclassifying negative samples as positive.

\section{Conclusion}\label{conclusion}
In this research, we have applied several ML models to predict whether a patient will have coronary heart disease in the next 10 years. It was found that the predictor variable was severly imbalanced and thus, we employed two class balancing techniques: Undersampling and oversampling, to balance the class. In terms of accuracy, the random forest model, with random oversampling, was best performing, having an accuracy of 0.84. However, from the ROC and precision-recall curve, we observe that the model's performance was far from optimal and it leave a lot of room for improvement in reducing the number of false positives. In addition, we would like to point out that most, if not all, the other studies in the literature also suffer from this ``FPR trap'', but they do not report this as a drawback of their method, however, we have chosen to do so here.

In future considerations, we will factor in the latter consideration and try and apply more sophisticated models to obtain higher accuracy and a reduced FPR. This work serves to set the tone for our future studies, and thus is a benchmarking paper. 

\section*{Declarations}

\begin{itemize}
\item \textbf{Funding:} J.A.K. and M.A.Z. acknowledge that this research is supported by grant number 23070, provided by Zayed University and the government of the UAE.
\item \textbf{Conflict of interest/Competing interests:} The authors declare that there are no conflicts of interest. 
\item \textbf{Ethics approval and consent to participate:} None required.
\item \textbf{Consent for publication:} The authors grant full consent to the journal to publish this article.
\item \textbf{Data availability:} \url{https://www.kaggle.com/datasets/aasheesh200/framingham-heart-study-dataset}.
\item \textbf{Materials availability:} N/A
\item \textbf{Code availability:} Available on a reasonable request from the corresponding author. 
\item \textbf{Author contribution:} All authors have contributed equally to this research.
\end{itemize}


\end{document}